\documentclass{article} 
\usepackage[final]{colm2026_conference}

\usepackage[protrusion=true,expansion=true,tracking=true,kerning=true]{microtype}
\usepackage{hyperref}
\usepackage{url}
\usepackage{booktabs}

\usepackage{lineno}
\usepackage[utf8]{inputenc} 
\usepackage[T1]{fontenc}    
\usepackage{hyperref}       
\usepackage{url}            
\usepackage{booktabs}       
\usepackage{amsfonts}       
\usepackage{nicefrac}       
\usepackage{xcolor}         

\usepackage{graphicx}
\usepackage{subcaption}
\usepackage{amsmath}
\usepackage{amssymb}
\usepackage{mathtools}
\usepackage{amsthm}
\usepackage[capitalize,noabbrev]{cleveref}
\usepackage{caption}

\definecolor{darkblue}{rgb}{0, 0, 0.5}
\hypersetup{colorlinks=true, citecolor=darkblue, linkcolor=darkblue, urlcolor=darkblue}

\title{Dispatch Overhead, Not Kernel Quality: The Performance Bottleneck in Single-Stream WebGPU LLM Inference}

\author{Jędrzej Maczan \\
Independent Researcher \\
Poland \\
\texttt{jedrzej@maczan.pl} \\
}

\begin{document}

\ifcolmsubmission
\linenumbers
\fi

\maketitle

\begin{abstract}
Large Language Models are increasingly deployed in resource-constrained, single-stream settings, from internet browsers to edge devices, where requests arrive one at a time at batch size 1 rather than in large batches. In this regime, where WebGPU serves as a modern cross-platform standard, per-operation dispatch overhead, rather than kernel compute, can dominate inference cost, yet it remains poorly characterized. 

In this work, we introduce a sequential-dispatch measurement method and show that naive single-operation measurements overestimate per-dispatch cost by conflating dispatch with synchronization. Using our method, we measure the per-dispatch cost and show that it is independent of data type used. Through a controlled fusion experiment, we show that the dispatch overhead, not kernel quality, is the bottleneck at batch size 1, and isolate the dispatch count as the cause. 

We conclude that at batch size 1, the effective approach to LLM inference optimization in WebGPU is reducing dispatch count. Our findings point to dispatch amortization, in the inference engines and in the WebGPU specification, as a path to practical browser-based inference.
\end{abstract}

\section{Introduction}

Large Language Models (LLMs) are increasingly deployed outside the datacenters, like in web browsers~\citep{mlcai2024webllm} and on the edge~\citep{cai2026edgesurvey, kodavanti2026unlockingedgedeploymentondevice}. WebGPU has emerged as the cross-platform standard enabling this, exposing a common interface across diverse hardware. Performance engineering for these systems has followed familiar paths: quantization~\citep{frantar2023gptq, lin2024awq, dettmers2022llmint8}, kernel optimization~\citep{dao2022flashattention, hidaka2024webgpublas, nuss2024webgpumatmul}, and memory efficiency~\citep{kwon2023pagedattention}. To our knowledge, however, no study characterizes the cost of WebGPU \emph{operation dispatch} in the context of machine learning inference.

WebGPU-based inference is hard in ways native execution is not. WebGPU code, like any code running in a browser, must meet high security standards that clash with the need for raw performance: every operation is validated before execution~\citep{w3c2024webgpu}. The execution environment is sandboxed and cannot be controlled by the inference engine, because it inherits whatever limits the browser imposes. And because the engine cannot reason about the target hardware, kernel (WGSL shader~\citep{w3c2024wgsl}) hyperparameters must be chosen from a ``safe'' set of general heuristics rather than tuned per device, as in LlamaWeb~\citep{levine2026llamaswebmemoryefficientperformanceportable}. Factors like these give rise to performance challenges distinct from native inference, and require new ways of measuring performance.

We find that the wall-clock measurements of operation performance overestimate per-dispatch cost by $\sim$20x, by conflating dispatch with synchronization. Each operation, when measured in isolation, carries a significant GPU-CPU synchronization cost that real autoregressive decoding incurs only once per token, not once per operation. Concretely, we introduce a \textbf{sequential-dispatch measurement method}: $N$ operations are dispatched together and synchronized only once at the end, which isolates the true cost of a single dispatch. On Dawn~\citep{google2024dawn}, the 497~$\mu$s single-operation measurement is dominated by synchronization (450~$\mu$s), with 24--36~$\mu$s dispatch cost on Vulkan and 32--71~$\mu$s on Metal. These times are identical for float32 and float16.

Knowing how to measure dispatch cost, we ask how much does dispatch actually affect performance? We test this in the area of kernel fusion, which ~\citet{levine2026llamaswebmemoryefficientperformanceportable} explicitly left for the future work. We find that fused kernels, using the \emph{same} WGSL shaders and saving a negligible amount of memory traffic, improve throughput by 53\% while cutting dispatches from 876 to 564. That fusion improves throughput is well known~\citep{snider2023xla, shi2023welder}. Our contribution is the \emph{controlled isolation} of dispatch count as the cause, and ruling out the kernels (shaders) quality and memory-bandwidth explanations that fusion is usually credited with. At batch size 1, the effective approach to LLM inference optimization in WebGPU is to reduce dispatch count, not to optimize kernels~\citep{mirage2025mpk, dong2026adamk}.

 \begin{figure}[t]
  \centering
  \includegraphics[width=0.8\linewidth, height=0.25\textheight, keepaspectratio]{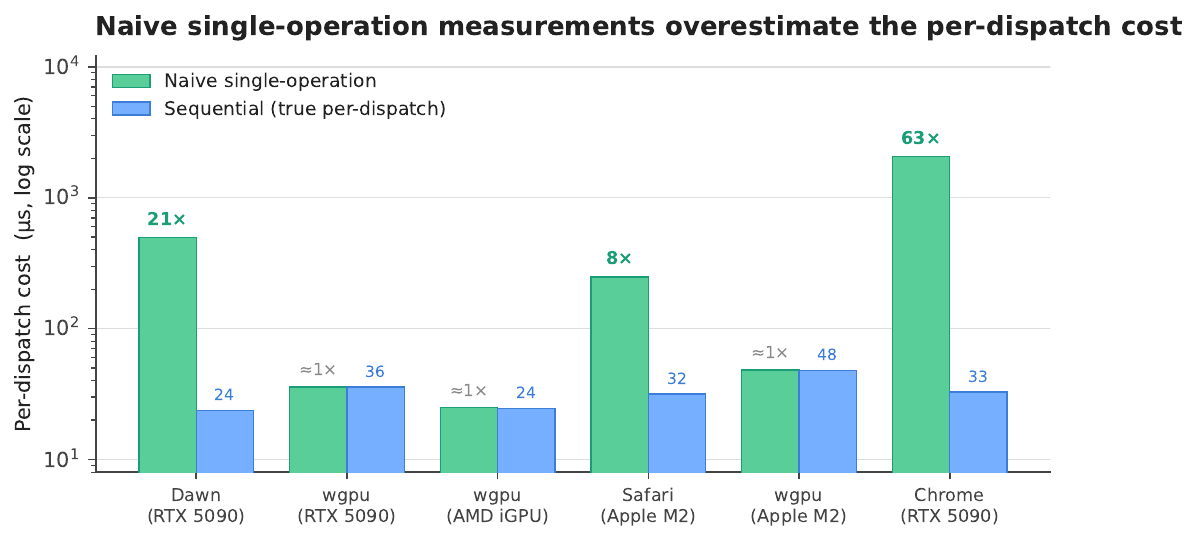}
  \caption{Naive single-operation measurement overestimates per-dispatch cost by conflating dispatch with synchronization. The figure compares the single-operation measurement with our sequential measurement, which amortizes that one-time synchronization away.}
  \label{fig:overestimate}
\end{figure}

\section{Related Work}

\textbf{Browser-based LLM inference.} The adoption of WebGPU and the rise of LLMs has produced several inference engines targeting browsers, including WebLLM~\citep{mlcai2024webllm}, Transformers.js~\citep{huggingface2024transformersjs}, and most recently LlamaWeb. The engines focus on quantization, kernels and their tradeoff between performance and portability, and memory efficiency. \citet{levine2026llamaswebmemoryefficientperformanceportable} reports state-of-the-art decode throughput and provides an extensive study on multiple kinds of hardware. They explicitly leave kernel fusion to reduce dispatch overhead as future work, which is the gap we address in our experiment. WeInfer~\citep{chen2025weinfer} targets the synchronization stage by postponing GPU-CPU read \emph{across} tokens, while our work characterizes and reduces the dispatch overhead \emph{within} each forward pass. The synchronization WeInfer optimizes is the one our sequential method amortizes. Thus, it makes our two studies complementary. Unlike WeInfer, which reports the aggregated speedups, we isolate dispatch count as the bottleneck through a controlled experiment.

\begin{figure}[t]
\centering
\includegraphics[width=0.8\linewidth, height=0.25\textheight, keepaspectratio]{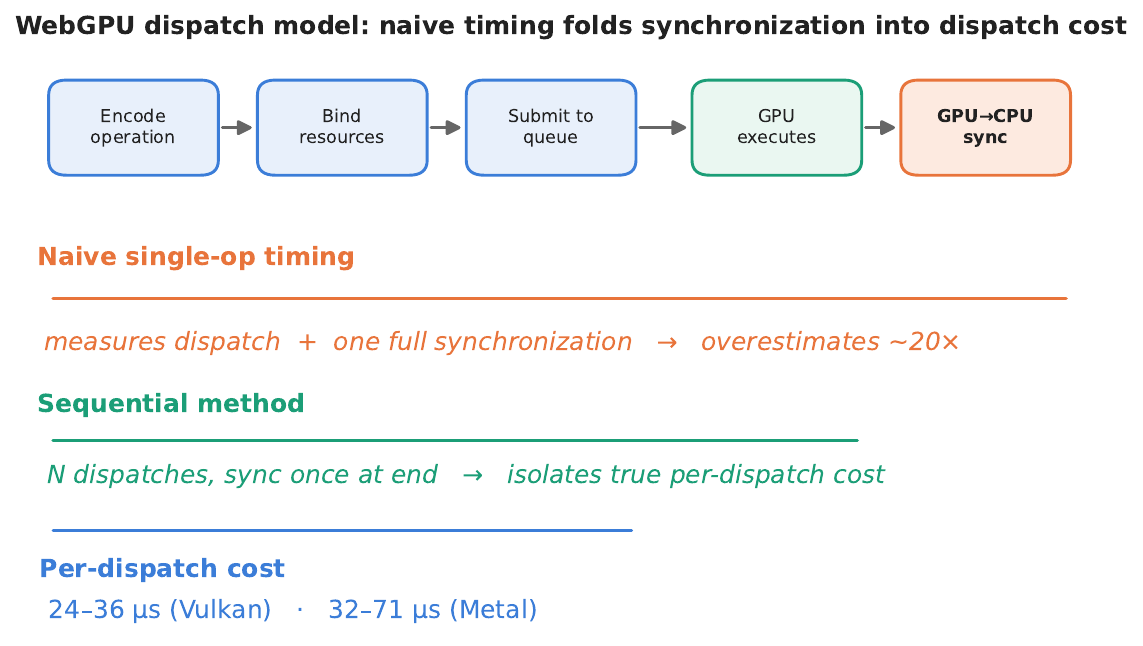}
\caption{The WebGPU dispatch model and the two measurement methods. In WebGPU, each
operation is encoded, has its resources bound, and is submitted to the queue, after which the GPU executes and a GPU-CPU synchronization returns the results. Naive single-operation timing spans one encode-through-synchronize round trip and folds the one-time synchronization into the per-dispatch measurement, overestimating the results by up to $\sim$20$\times$. Our sequential method issues $N$ dispatches and synchronizes only once at the end, isolating the true marginal cost of a single dispatch.}
\label{fig:model}
\end{figure}

\textbf{Dispatch and launch overhead.} On native APIs, kernel launch overhead is well studied: for example a latency of CUDA kernels launch is a few microseconds and can be amortized further with CUDA Graphs~\citep{nvidia2024cudagraphs}. Vulkan's design~\citep{khronos2024vulkan} tries to reduce the CPU overhead relative to OpenGL. A recent work quantifies its impact on LLM inference~\citep{maczan2026characterizingwebgpudispatchoverhead}. \citet{vellaisamy2025characterizing} characterize kernel launch overhead on GPU-CPU line and find that systems remain CPU-bound at low batch sizes. \citet{dong2026adamk} reports that launch overhead alone can account for $14.6\%$ of end-to-end decode latency in production serving. We provide the analogous characterization for WebGPU. For the reference, in Figure~\ref{fig:model} we present WebGPU dispatch model overview.

\textbf{Kernel fusion and megakernels.} In deep learning compilers and inference engines, operation/kernel fusion is a standard technique for reducing memory traffic and launch overhead~\citep{dao2022flashattention,chen2018tvm, snider2023xla,shi2023welder}. Pushed to its limit, the megakernels line of work~\citep{eventtensor2026} fuses the entire into a single kernel to eliminate launch overhead~\citep{mirage2025mpk,hazy2025megakernel,dong2026adamk}. \citet{vellaisamy2025characterizing} shows the potential of this method to mitigate the bottleneck in LLM inference specifically. In our study, we use fusion as an instrument in isolating dispatch count as the cause of the observed throughput gains.
\section{Background}

WebGPU is the modern cross-platform successor to graphics standards like OpenGL and WebGL, providing a common interface to the major native APIs (Vulkan, Metal, Direct3D) across all major browsers and as standalone native implementations - Dawn~\citep{google2024dawn} and wgpu~\citep{gfxrs2024wgpu}. Unlike the graph-submission model of WebNN~\citep{w3c2026webnn}, which amortizes per-operation dispatch by lowering a whole network graph to an OS-level runtime, the WebGPU API does not. WebGPU follows a deferred command-buffer model. Operations are encoded into command buffers, connected using bind groups, submitted to a queue, and synchronized with explicit waits or buffer mappings. Each dispatch adds an overhead from encoder creation, bind group setup, and queue submission. Autoregressive LLM decoding at batch size 1 is built from hundreds of such small dispatches per token, with a required GPU-CPU synchronization for token selection. In such setup, per-dispatch overhead is a significant factor of LLM inference performance.

 \begin{figure}[t]
\centering
\includegraphics[width=0.8\linewidth, height=0.25\textheight, keepaspectratio]{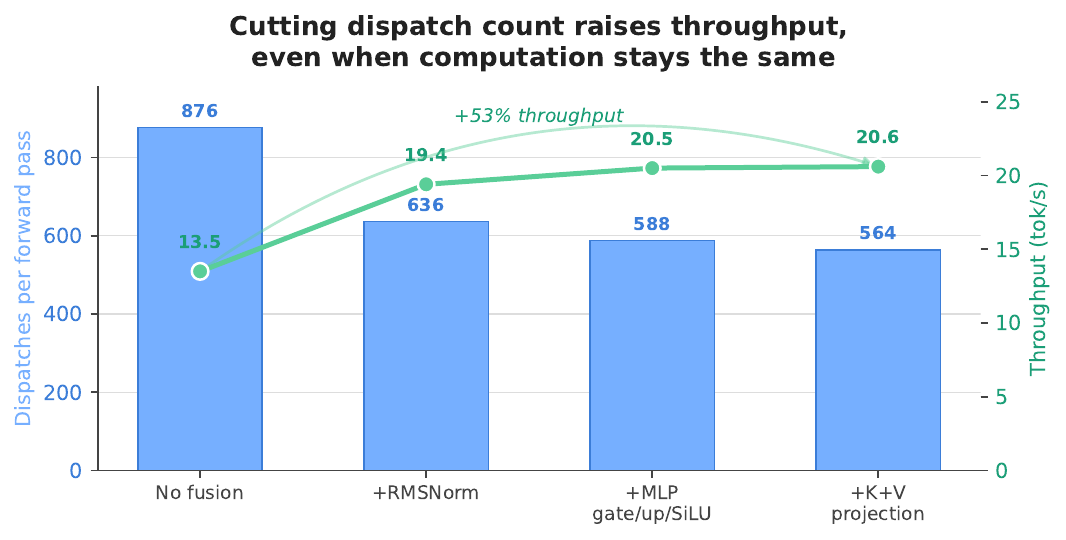}
\caption{Our controlled fusion experiment isolates dispatch count and identifies it as the performance lever.}
\label{fig:fusion}
\end{figure}
\section{Method}
\label{sec:method}
We make all the measurements using \emph{torch-webgpu}~\citep{Maczan_torch-webgpu_2025}, an open-source out-of-tree PrivateUse1-based~\citep{pytorch_privateuse1} PyTorch~\citep{ansel2024pytorch2} backend and compiler for WebGPU. It compiles models to WGSL shaders and executes them with Dawn. Its simplicity makes per-dispatch measurements and fusion experiments straightforward and reproducible by the research community.

\subsection{Sequential-dispatch measurement}

The naive way to measure dispatch cost is to time a single operation end to end, from encode, through resources binding, submit, actual computation to a synchronization. This measurement conflates two distinct costs of the dispatch itself and a full GPU-CPU synchronization, and attributes both to the dispatch. In real decoding, however, synchronization happens only once per token. Our sequential-dispatch method mirrors this behavior. We submit $N$ identical dispatches with a single synchronization at the end and then divide the total time by $N$. The one-time synchronization is amortized and leaves the true marginal cost of one dispatch. We apply this across four GPU vendors (NVIDIA, AMD, Apple, Intel) and multiple WebGPU implementations (Dawn, wgpu-native), on Vulkan and Metal backends.

\subsection{Measuring the performance impact of dispatch}

To test whether dispatch overhead, not kernel quality, limits throughput at batch size 1, we run a controlled kernel fusion experiment. We fuse groups of operations so the dispatch count per forward pass drops from 876 to 564, while keeping the underlying computation unchanged. Because the WGSL shaders are essentially unchanged, any throughput difference cannot be attributed to improved kernel quality. Similarly, fusion eliminates only a small amount of intermediate memory traffic (about 1.8~MB, under 1 $\mu$s), so the difference cannot be attributed to reduced memory traffic. The only variable that changes meaningfully is the number of dispatches. We report this experiment on our reference setup (NVIDIA RTX 5090 32~GB, Dawn, Vulkan, float32).

\subsection{Characteristics of WebGPU dispatch}
\begin{figure}[t]
  \centering
  \includegraphics[width=0.8\linewidth, height=0.25\textheight, keepaspectratio]{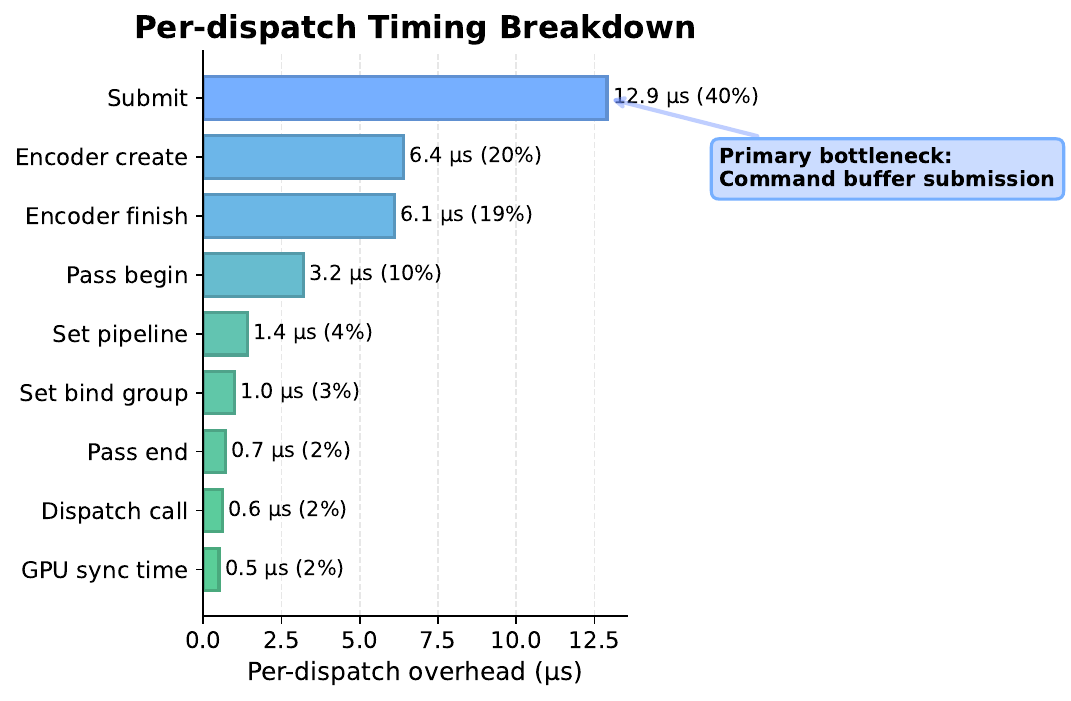}
  \caption{Per-dispatch overhead breakdown. Command buffer submission (Submit) is the dominant source of overhead - taking about 40\% of total CPU time - 12.9 µs per dispatch. Encoder creation and completion contribute an additional 12.5 µs combined, while pipeline setup, bind-group binding, and dispatch invocation are a comparatively small costs (<1.5 µs each). We identify command buffer submission as the primary bottleneck. This result supports our claim that reducing dispatch count is more impactful than further optimizing pipeline or binding operations.}
  \label{fig:breakdown}
\end{figure}

To understand the characteristics of dispatch itself, we measured different steps of WGSL shaders dispatch. In figure~\ref{fig:breakdown} we present the breakdown of averaged measurements of operations specific to WebGPU during dispatch. We find it informative in context of future decisions regarding WebGPU API design and its implementations.

\section{Results}

\subsection{True per-dispatch cost is far below naive estimates}

Figure~\ref{fig:overestimate} compares single-operation and sequential measurements. Single-operation measurements overestimate per-dispatch cost by roughly 20x on native Dawn and more on Chrome and Safari, since it folds in a 450~$\mu$s synchronization
cost that decoding pays only once per token. The sequentially measured cost is 24--36~$\mu$s on Vulkan and 32--71~$\mu$s on Metal, and is consistent across four GPU vendors on the same backend. Because a dispatch cost reflects command encoding and submiss
ion rather than the arithmetic the kernel performs, it is independent of data type (identical for float32 and float16), and we attribute it to the WebGPU API.

\subsection{Dispatch count, not kernel quality, is the bottleneck}

Figure~\ref{fig:fusion} shows the controlled fusion experiment results. Reducing dispatches from 876 to 564, with the same WGSL shaders and negligible memory savings, improves throughput by 53\% (Time to First Token (TTFT) drops from 71.4~ms to 41.6~ms). Since kernel quality and memory traffic are held constant by construction, we attribute the improvement to the 312 fewer dispatches. This isolates per-operation dispatch overhead, not kernel quality, as the bottleneck at batch size 1. This result is consistent with the broader observation that highly optimized individual kernels do not proportionally improve end-to-end throughput at batch size 1 - when per-operation overhead dominates, kernel speed is not the binding constraint.

\section{Discussion}

The conventional performance optimization levers, such as writing better kernels, quantization, improving the memory layout, are not sufficient in case of inference using WebGPU. The binding constraint is the number of dispatches and the per-operation overhead added by each of them. The practical implication is that fusion, which reduces dispatch count, even if without any other additional computation improvements, is among the high value WebGPU-specific optimizations available. With this finding, we fill the missing kernel fusion research impact for WebGPU, left for the future work by \citet{levine2026llamaswebmemoryefficientperformanceportable}. Because per-dispatch cost is a property of the WebGPU API rather than of any particular model or shader, the deeper improvements might require changes in WebGPU specification itself. Dispatch amortization, mechanisms analogous to command-graph capture and replay (as in CUDA Graphs), would help avoid re-validating and re-submitting every operation. We recognize that such changes are non-trivial and have to be weighed against the browser security model, that motivates per-operation validation in the first place.

\noindent\textbf{Scope.} The causal fusion experiment, on a single reference platform (RTX 5090, Dawn, Vulkan) at batch size 1 with float32, should be read as evidence on this platform and regime rather than a universal claim. Batch sizes greater than one, where kernel compute begins to dominate, are left for future work.

\bibliography{colm2026_conference}
\bibliographystyle{colm2026_conference}


\end{document}